\documentclass[lettersize,journal]{IEEEtran}
\usepackage{graphicx}
\usepackage{lipsum} 
\usepackage{soul}
\usepackage{xcolor}
\sethlcolor{yellow}
\usepackage{hyperref}
\usepackage{amsmath}
\usepackage{amssymb}
\usepackage{booktabs}
\usepackage{subcaption}
\usepackage{tikz} 

\begin{document}
\title{Heteroscedastic Neural Networks for Path Loss Prediction with Link-Specific Uncertainty}

\author{Jonathan~Ethier
\thanks{This work has been submitted the IEEE Antennas and Wireless Propagation Letters (AWPL) journal, December 2025. J. Ethier is with the Communications Research Centre, Canada. e-mail: jonathan.ethier@ised-isde.gc.ca}}
\maketitle

\begin{abstract}
Traditional and modern machine learning–based path loss models typically assume a constant prediction variance. We propose a neural network that jointly predicts the mean and link-specific variance by minimizing a Gaussian negative log-likelihood, enabling heteroscedastic uncertainty estimates. We compare shared, partially shared, and independent-parameter architectures using accuracy, calibration, and sharpness metrics on blind test sets from large public RF drive-test datasets. The shared-parameter architecture performs best, achieving an RMSE of 7.4 dB, 95.1\% coverage for 95\% prediction intervals, and a mean interval width of 29.6 dB. These uncertainty estimates further support link-specific coverage margins, improve RF-planning and interference analyses, and provide effective self-diagnostics of model weaknesses.
\end{abstract}


\section{Introduction}

\IEEEPARstart{A}{ccurate} path loss modeling is critical for coverage planning and interference management in modern wireless systems. Planners typically use deterministic path loss predictions for basic planning and, to mitigate risk, add margins to account for outage probability and capacity provisioning. Thus, it is advantageous to predict not only a point estimate but also a distribution that captures variability arising from operating frequency, link distance, and obstructions. This motivates the development of probabilistic path loss models that can explicitly represent and quantify such uncertainty.

Classical models such as COST-231 \cite{COST231FinalReport1999}, {WINNER} II \cite{WINNERII2007}, and 3GPP UMa NLOS Model \cite{3gpp38901} predict path loss using log-distance or physical models and assume a fixed variance of error (homoscedastic uncertainty). However, real-world uncertainty is not constant (it is heteroscedastic).

Recent advances in machine learning (ML) for path loss prediction show promise in accuracy, computational efficiency, and model generalization \cite{deeplearning_pathloss_transactions,ethierFeatures,bocusPaper}. In all these cases, optimization relies on the mean squared error (MSE). Using MSE for point predictions implicitly assumes homoscedastic Gaussian residuals, mirroring the limitations of classical models.

Path loss uncertainty is inherently link-dependent, resulting in heteroscedasticity. This motivates our exploration of Gaussian negative log-likelihood (NLL) \cite{hastie2009elements} as an alternative loss function for path loss modeling. Under a Gaussian error model, NLL enables estimation of both the mean and variance of the distribution, thereby capturing link-dependent uncertainty rather than assuming a fixed variance.

While the authors in \cite{momentmethoduncertainty} use NLL to quantify uncertainty in numerical approximations of closed-form operators, we instead apply NLL to data-driven, heteroscedastic path loss prediction. To our knowledge, this is the first systematic use of NLL in this setting, with rigorous blind testing across diverse environments. Alternatively, estimates of uncertainty can be acquired post-hoc and distribution-free using a pre-existing point-predictive model and additional calibration data \cite{bose2025uncertaintyestimationpathloss}; however, this is distinct from the model-intrinsic and rigorous blind testing approach being explored here.

In this work, we make three main contributions. First, we introduce an ML-based path loss model using the Gaussian NLL as the loss function, providing path loss distributions through the joint predictions of mean and variance. Second, we provide rigorous model validation and investigate shared, partially shared, and independent parameter architectures and identify which approach is most effective. Last, we show the value of heteroscedasticity by providing link-dependent uncertainty that enhances real-world deployments and exposes model weaknesses for targeted improvements.

\section{Methodology}
\label{sec:methodology}
\subsection{Problem Formulation}
In supervised machine learning, models are optimized by minimizing a loss function that measures the discrepancy between predictions and ground truth \cite{Goodfellow-et-al-2016}. One of the most common loss functions in regression problems is MSE, as defined in \eqref{mse}.

\begin{equation}
\mathcal{L}_{\mathrm{MSE}} = \frac{1}{N} \sum_{i=1}^{N} \left(y_i - \hat{y}_i\right)^2
\label{mse}
\end{equation}
where $y_i$ represents the ground truth and $\hat{y}_i$ is the model prediction. To obtain heteroscedastic estimates (non-constant variance across inputs) of prediction uncertainty, we require a loss function  using both mean and variance as optimizable parameters, namely NLL, as defined in \eqref{nll}.

\begin{equation}
\mathcal{L}_{\text{NLL}} = \frac{1}{N} \sum_{i=1}^{N} \left[ \frac{1}{2} \log(2\pi\sigma_i^2) + \frac{(y_i - {\mu}_i)^2}{2\sigma_i^2} \right]
\label{nll}
\end{equation}
where $\mu_i$ and $\sigma_i^2$ are the predicted mean and variance for sample $i$, respectively. The assumption when using NLL as a loss function is that the error statistics follow a Normal distribution. This is a reasonable assumption in practice due to the aggregation of many independent error sources naturally converging to a Gaussian distribution. Note that $\mu_i$ serves as the predicted mean of the distribution (analogous to $\hat{y}_i$ in MSE), while $\sigma_i^2$ quantifies the prediction-specific uncertainty. 


\subsection{Architecture Variants}
\label{sec:architectures}
Consider a functional transformation of the input feature vector 
$\mathbf{x}_i$ using model parameters $\boldsymbol{\theta}_\mu$ and 
$\boldsymbol{\theta}_\sigma$ to predict the mean $\mu_i$ and variance 
$\sigma_i^2$ from \eqref{nll}, as shown in \eqref{fmu} and \eqref{fsig}, respectively.
\begin{equation}
\mu_i = f_\mu(\mathbf{x}_i; \boldsymbol{\theta}_\mu)
\label{fmu}
\end{equation}
\begin{equation}
\sigma^2_i = f_{\sigma}(\mathbf{x}_i; \boldsymbol{\theta}_{\sigma}) \\
\label{fsig}
\end{equation}

In this work, the functions $f_\mu$ and $f_\sigma$ are implemented as dense neural networks, though the general framework presented here applies to other architectures. The model hyperparameters (number of layers and neurons, choice of activation functions, etc.) can be organized in three ways:
\begin{align}
\text{Shared} &: \quad \boldsymbol{\theta}_\mu = \boldsymbol{\theta}_\sigma \\
\text{Partial} &: \quad \boldsymbol{\theta}_\mu \cap \boldsymbol{\theta}_\sigma \neq \emptyset \;\text{and}\; \boldsymbol{\theta}_\mu \ne \boldsymbol{\theta}_\sigma\\
\text{Independent} &: \quad \boldsymbol{\theta}_\mu \cap \boldsymbol{\theta}_\sigma = \emptyset
\end{align}

These configurations correspond to: (a) a single network predicting both outputs (shared parameters), (b) a network with shared internal features feeding two independent outputs, i.e., separate output layers (partially-shared parameters), and (c) two entirely separate networks (independent parameters), with all three architectures trained on identical data and features.
The motivation for investigating these networks is threefold: (1) determining if there are any advantages to jointly estimating mean and variance versus independent estimation, (2) identifying whether performance metrics are architecture dependent, and (3) evaluating potential training stability issues associated with each architecture.



\subsection{Performance Metrics}
NLL provides a reliable assessment for probabilistic predictions by evaluating how much probability the model assigns to the observed outcomes, providing a single score that penalizes both inaccurate predictions and miscalibrated uncertainty (i.e., over- or under-confidence). However, as a single metric, NLL risks obscuring whether poor performance is due to inaccurate predictions, poorly calibrated uncertainty estimates, or a combination of both. To address this, we decompose the evaluation into three key metrics, starting with RMSE (square root of Eq. \eqref{mse}), which provides an interpretable assessment of prediction accuracy. Furthermore, assessing uncertainty estimates requires two equally important metrics: calibration (how well predicted probabilities match observed frequencies) and sharpness (how tight the predictions are) \cite{picp_and_mpiw_paper}. 

For calibration, we use Prediction Interval Coverage Probability (PICP) in \eqref{picp}. For sharpness, we use Mean Prediction Interval Width (MPIW) in \eqref{mpiw}. In both equations, $y_i$ represents the ground truth; $\hat{y}_i^L$ and $\hat{y}_i^U$ are the lower and upper bounds of model predictions; N is the total number of samples; and $I(\cdot)$ is the indicator function (1 if true, 0 otherwise).

\begin{equation}
\text{PICP} = \frac{1}{N} \sum_{i=1}^{N} \mathbb{I}\left(\hat{y}_i^{L} \leq y_i \leq \hat{y}_i^{U}\right)
\label{picp}
\end{equation}

\begin{equation}
\text{MPIW} = \frac{1}{N} \sum_{i=1}^{N} (\hat{y}_i^U - \hat{y}_i^L)
\label{mpiw}
\end{equation}

To compute PICP and MPIW, we need to construct a prediction interval for each sample, defined by the lower $\hat{y}_i^L$ and upper $\hat{y}_i^U$ boundaries. For a Gaussian distribution with predicted mean $\mu_i$ and variance $\sigma_i^2$, these bounds are computed as $\hat{y}_i^L = \mu_i - z_{\alpha/2}\sigma_i$ and $\hat{y}_i^U = \mu_i + z_{\alpha/2}\sigma_i$, where $z_{\alpha/2}$ is the critical value corresponding to the desired confidence level ($z_{0.025} = 1.96$ for 95\% coverage). 

For model assessment, we start by constructing a prediction interval intended to cover 95\% of the ground truth values. We then use PICP to evaluate whether the actual coverage matches this level (calibration). Given good calibration (PICP $\approx$ 95\%), MPIW can be used to assess the tightness (sharpness) of the intervals. The ideal model will be accurate (low RMSE), have good calibration (PICP close to desired probability) and sharp intervals (low MPIW). Wireless coverage and interference studies typically focus on 95\% intervals \cite{95percconfidence}, which is the focus of this work.


\subsection{Dataset Description, Preprocessing, and Model Features}
\label{sec:data}
We use the same training data as \cite{ethierFeatures} and \cite{bocusPaper}: RF drive test data acquired by the UK's Office of Communications (Ofcom) \cite{OFCOM_DATA_OPEN}. Path profiles, consisting of obstruction heights above sea level while accounting for Earth curvature, are extracted from UK Open Data \cite{UK_DTM_DSM}, using code from \cite{SAFE_GITHUB}. Model features are derived from these path profiles by using samples above the measurement noise floor with a 6~dB margin, as suggested in \cite{OFCOM_DATA_OPEN}. A total of 20\,000 samples were extracted (without replacement, at random) for each of the six measurement frequencies (449, 915, 1802, 2695, 3602, and 5850 MHz) within each of the six drive tests (London, Merthyr Tydfil, Nottingham, Southampton, Stevenage, Boston), resulting in a training set size of 720\,000 samples (20\,000 samples $\times$ 6 frequencies $\times$ 6 drive test locations).

Three model features are used in this study: frequency (f), link distance (d) and total obstruction depth (o). The first two are obtained from the datasets \cite{OFCOM_DATA_OPEN}, while the total obstruction depth (defined in \cite{ethierFeatures}) is acquired from UK Open Data \cite{UK_DTM_DSM}. These three features form a strong foundation for baseline model performance, as they provide frequency and distance (fundamental to any path loss model), along with a single informative feature associated with obstructions. While additional features (e.g., terrain roughness, clutter type) could improve performance, this minimal feature set highlights NLL’s fundamental benefits for uncertainty quantification and preserves computational efficiency.

\section{Architecture Performance Analysis}
\label{sec:architecture}

\subsection{Model Optimization Strategy}
\label{subsection:optstrat}
For all three architectures, we use leave-one-group-out cross-validation (LOGO-CV) with the six drive tests from Section \ref{sec:data}. Given the varied distribution of environments, these holdouts form a comprehensive set of tests to assess model generalization. In each of the six folds, the training portion (five drive tests) is randomly divided into 80\% for training and 20\% for validation. Each split is repeated 10 times to capture variations due to random initialization and train/validation partitioning. Training uses a batch size of 1024, ReLU activations, 25\% dropout after each layer in training-only, Adam optimizer with a learning rate of 0.01, and early stopping at 100 epochs. For fair comparison among the architectures, each network should have a similar number of optimizable parameters ($\sim$ 4\,500) and network depth (2), with the following three configurations:
\begin{itemize}
    \item \textbf{Shared}: single network, 2 hidden layers, 64 neurons per layer, 4\,546 parameters total.
    \item \textbf{Partial}: 1 hidden layer with 45 neurons, with internal features shared with each independent head, each having 1 hidden layer with 45 neurons, 4\,412 parameters total.
    \item \textbf{Independent}: each independent network has 2 hidden layers, 45 neurons each, 4\,592 parameters total.
\end{itemize}

Preliminary experiments with $>10\,000$ parameters and $>2$ layers showed worse validation scores, demonstrating overfitting. All three architectures have similar complexity; training averages 5.1 minutes per fold on a low-cost CPU instance, and inference exceeds 10\,000 predictions/s.


\subsection{Training Stability}

\begin{figure}[htbp]
    \centering
    \begin{subfigure}{\columnwidth}
        \centering
        \includegraphics[width=2.93in]{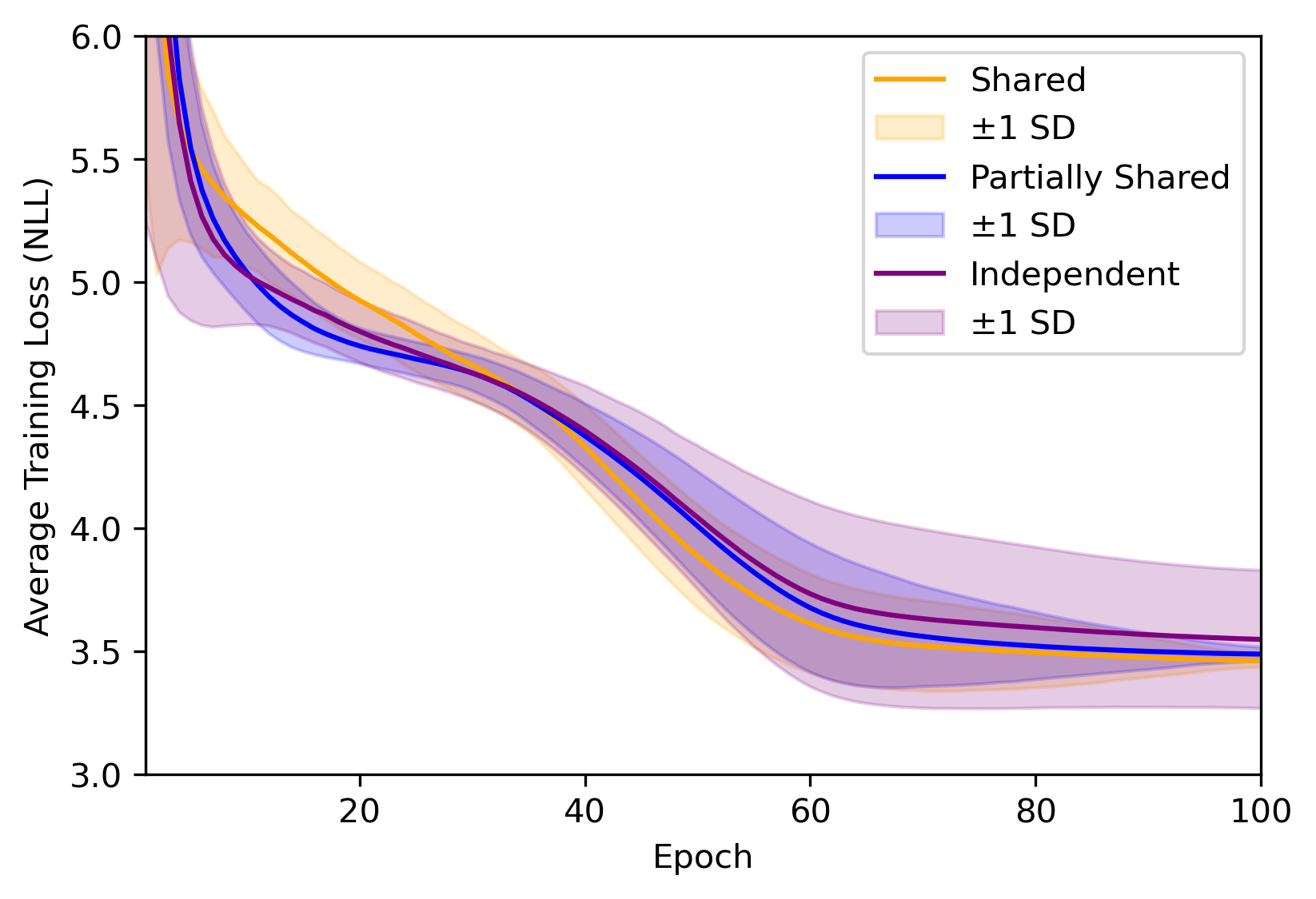}
        \caption{}
        \label{fig:sub1}
    \end{subfigure}
    \begin{subfigure}{\columnwidth}
        \centering
        \includegraphics[width=2.93in]{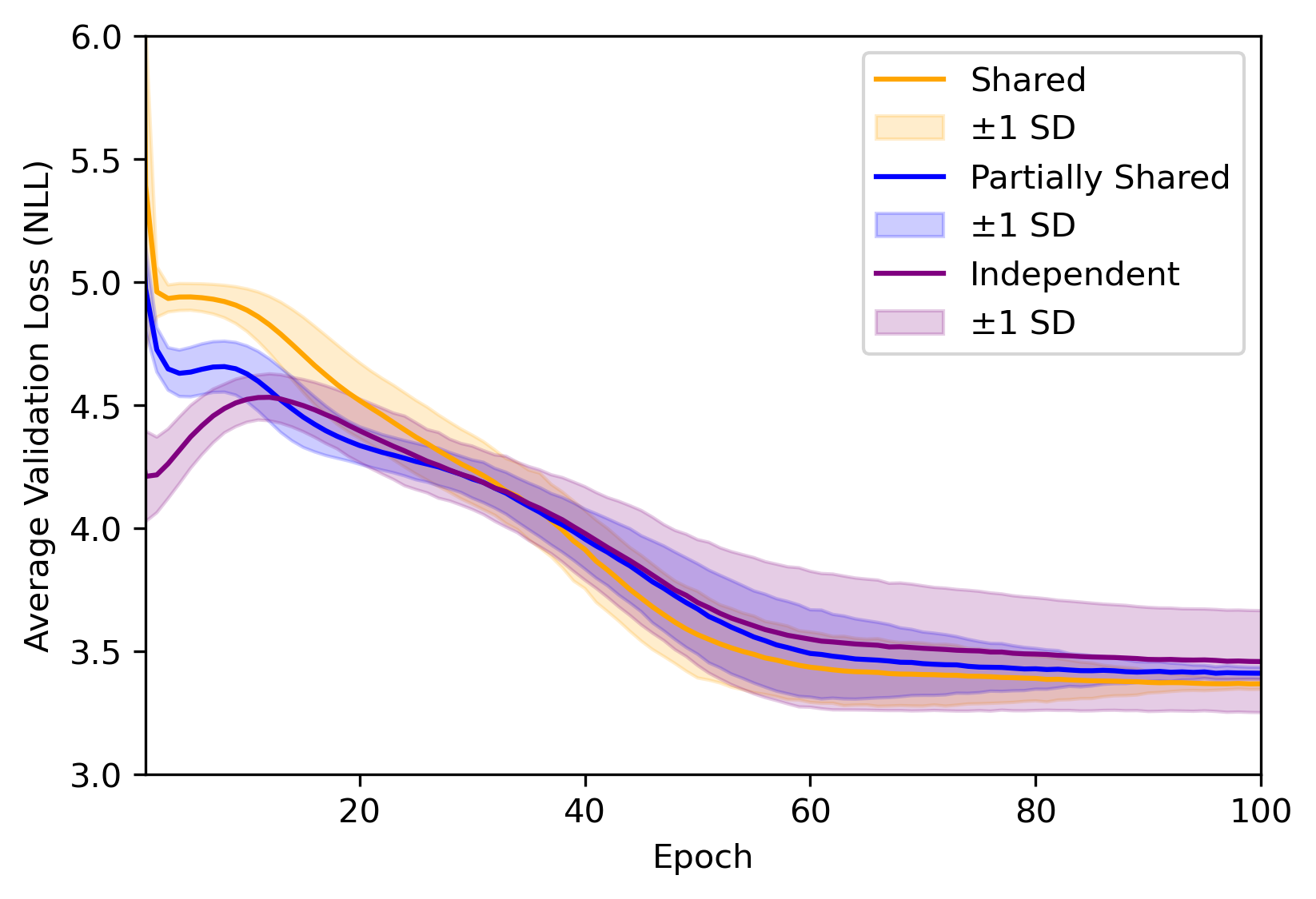}
        \caption{}
        \label{fig:sub1}
    \end{subfigure}
    \caption{NLL (a) training and (b) validation loss using different architectures}
    \label{fig:nll_convergence}
\end{figure}

Using the hyperparameters outlined in the previous section, we produce the training and validation plots shown in Fig.~\ref{fig:nll_convergence}, and Table~\ref{tab:nll_table} summarizes the converged NLL values. For these plots, we calculate the mean and SD of the loss curves across the 60 training scenarios (6 different holdouts $\times$ 10 runs each) described in Section~\ref{subsection:optstrat}. All three architectures converge to similar mean training and validation scores, with the shared-parameter architecture producing the lowest NLL, considering both the mean and SD of the results. The shared-parameter architecture's superior performance likely arises from implicit regularization; forcing the network to serve both objectives prevents overspecialization and encourages learning of more robust internal representations. 

\begin{table}[ht]
\centering
\caption{NLL scores for different architectures (lower is better; averaged over six training sets and 10 runs per set)}
\begin{tabular}{lcccc}
\toprule
 & \multicolumn{2}{c}{\textbf{Training}} & \multicolumn{2}{c}{\textbf{Validation}}\\
{\textbf{Architecture}} & Mean & SD & Mean & SD \\
\midrule
Shared      & 3.46 & 0.03 & 3.37 & 0.02 \\
Partial     & 3.49 & 0.03 & 3.41 & 0.03 \\
Independent & 3.55 & 0.28 & 3.46 & 0.21 \\
\bottomrule
\end{tabular}
\label{tab:nll_table}
\end{table}

\subsection{Accuracy, Calibration, and Sharpness on Blind Test Sets}
\label{subsection:testing}
We now consider the three key performance metrics on the blind test sets. The resulting accuracy, calibration, and sharpness for a 95\% prediction interval are shown in Table \ref{tab:data_table}, along with the SD of their scores across the 60 separate tests (6 LOGO-CV $\times$ 10 initializations each). The shared model architecture has the best mean scores and lowest SD for all three metrics. The PICP scores are close to the target 95\%, allowing us to confidently rely on the intervals provided by the three architectures. Paired t-tests on RMSE confirmed that the shared-parameter architecture significantly outperformed the partial architecture (p = 0.0013), which in turn significantly outperformed the independent architecture (p = 0.0218), with a Bonferroni-corrected threshold of  $\alpha = 0.025$ for two comparisons \cite{hair2010multivariate}. Similarly, paired t-tests conducted on PICP and MPIW metrics show that the shared-parameter architecture significantly outperforms the alternative models.

To provide a homoscedastic baseline, the accuracy, calibration, and sharpness provided by an MSE-optimized three-feature ML model \cite{ethierFeatures} is included, as shown in Table~\ref{tab:data_table}. The model uses identical features and similar architecture, and provides similar accuracy (RMSE 7.4 dB), sharper intervals (SD of 6.9 dB from validation residuals, resulting in a constant 27.0 dB MPIW), but the sharper intervals are the result of overconfidence with worse calibration (93.2\%). With no loss in accuracy, improved calibration, and an added advantage of link-specific heteroscedastic uncertainty, the use of NLL as a loss target offers only upsides in modeling utility. 

\begin{table}[ht]
\centering
\caption{Accuracy, calibration, and sharpness for blind test sets (averaged over six holdout tests and 10 runs per test)}
\begin{tabular}{lcccccc}
\toprule
 & \multicolumn{2}{c}{\textbf{RMSE [dB]}}
 & \multicolumn{2}{c}{\textbf{PICP$_{95}$}} 
 & \multicolumn{2}{c}{\textbf{MPIW$_{95}$ [dB]}} \\
\textbf{Architecture} & Mean & SD & Mean & SD & Mean & SD \\
\midrule
Shared                         & 7.4 & 0.1 & 95.1\% & 0.4\% & 29.6 & 0.4 \\
Partial                        & 7.7 & 0.2 & 94.7\% & 0.6\% & 30.0 & 0.6 \\
Independent                    & 8.3 & 1.5 & 95.0\% & 1.2\% & 32.5 & 6.1 \\
ML (MSE loss) \cite{ethierFeatures} & 7.4 & 0.7 & 93.2\% & 2.7\% & 27.0 & 0.0 \\
\bottomrule
\end{tabular}
\label{tab:data_table}
\end{table}

An additional homoscedastic baseline is provided with 3GPP UMa NLOS 38.901~\cite{3gpp38901}, which assumes a constant path loss SD of 6 dB for NLOS and 4 dB for LOS. Since UMa is crafted for urban environments, we limit the test holdout to London-only for a fair comparison. As the results in Table \ref{tab:data_table_london} confirm, the shared parameter model is similarly advantageous when blind tested in an urban environment, and is superior to the UMa model in terms of accuracy and calibration.
\begin{table}[ht]
\centering
\caption{Results for London-only blind tests (averaged over 10 runs)}
\begin{tabular}{lcccccc}
\toprule
 & \multicolumn{2}{c}{\textbf{RMSE [dB]}}
 & \multicolumn{2}{c}{\textbf{PICP$_{95}$}} 
 & \multicolumn{2}{c}{\textbf{MPIW$_{95}$ [dB]}} \\
\textbf{Architecture} & Mean & SD & Mean & SD & Mean & SD \\
\midrule
Shared                  & 7.1 & 0.1 & 95.2\% & 0.7\% & 28.5 & 0.6 \\
Partial                 & 7.2 & 0.2 & 96.3\% & 0.4\% & 29.6 & 0.5 \\
Independent             & 8.3 & 3.3 & 96.2\% & 1.8\% & 33.0 & 9.1 \\
UMa \cite{3gpp38901}    & 11.1 & 0.0 & 69.4\% & 0.0\% & 23.5 & 0.0 \\
\bottomrule
\end{tabular}
\label{tab:data_table_london}
\end{table}


\subsection{Assessment of Normality}
\label{subsection:normality}

We now assess whether the normality assumption underlying the NLL objective holds in our models. Given the large sample size ($>$100,000 test samples), we use kurtosis (tail heaviness) and skewness (asymmetry) as practical measures of Normality rather than tests like Shapiro-Wilk, which becomes overly sensitive to trivial deviations at this scale \cite{hair2010multivariate}. The mean and SD of these metrics (normalized by the predicted variance) are shown in Table \ref{tab:gauss_table} across the 60 tests for all three model architectures. Magnitudes of kurtosis and skewness $<1$ are commonly viewed as approximately Normal \cite{hair2010multivariate}. The shared-parameter architecture meets these criteria, with no model exceeding this threshold in any of the 10 runs. 

\begin{table}[ht]
\centering
\caption{Kurtosis and Skewness of Standardized Residuals}
\begin{tabular}{lcccccc}
\toprule
 & \multicolumn{3}{c}{\textbf{Kurtosis}} & \multicolumn{3}{c}{\textbf{Skewness}}\\
{\textbf{Architecture}} & Mean & SD & Max & Mean & SD & Max \\
\midrule
Shared      & 0.22 & 0.29 & 0.93 & 0.31 & 0.15 & 0.65 \\
Partial     & 0.18 & 0.16 & 1.54 & 0.25 & 0.12 & 0.73 \\
Independent & 0.38 & 0.87 & 5.45 & 0.29 & 0.23 & 1.28 \\
\bottomrule
\end{tabular}
\label{tab:gauss_table}
\end{table}


\subsection{Spatial Distribution of Predictive Uncertainty: A Use Case}
\label{subsection:usecase}

Path loss models are commonly used to generate spatial coverage maps from transmitters. Traditional approaches provide \textit{point predictions} with uniform uncertainty margins. Our method instead predicts a path loss distribution at each location, yielding spatially informed mean and SD values. This enables RF planners to apply link-specific margins—tighter or more relaxed as conditions warrant.

\begin{figure}[!ht]
    \centering
    \includegraphics[width=2.92in]{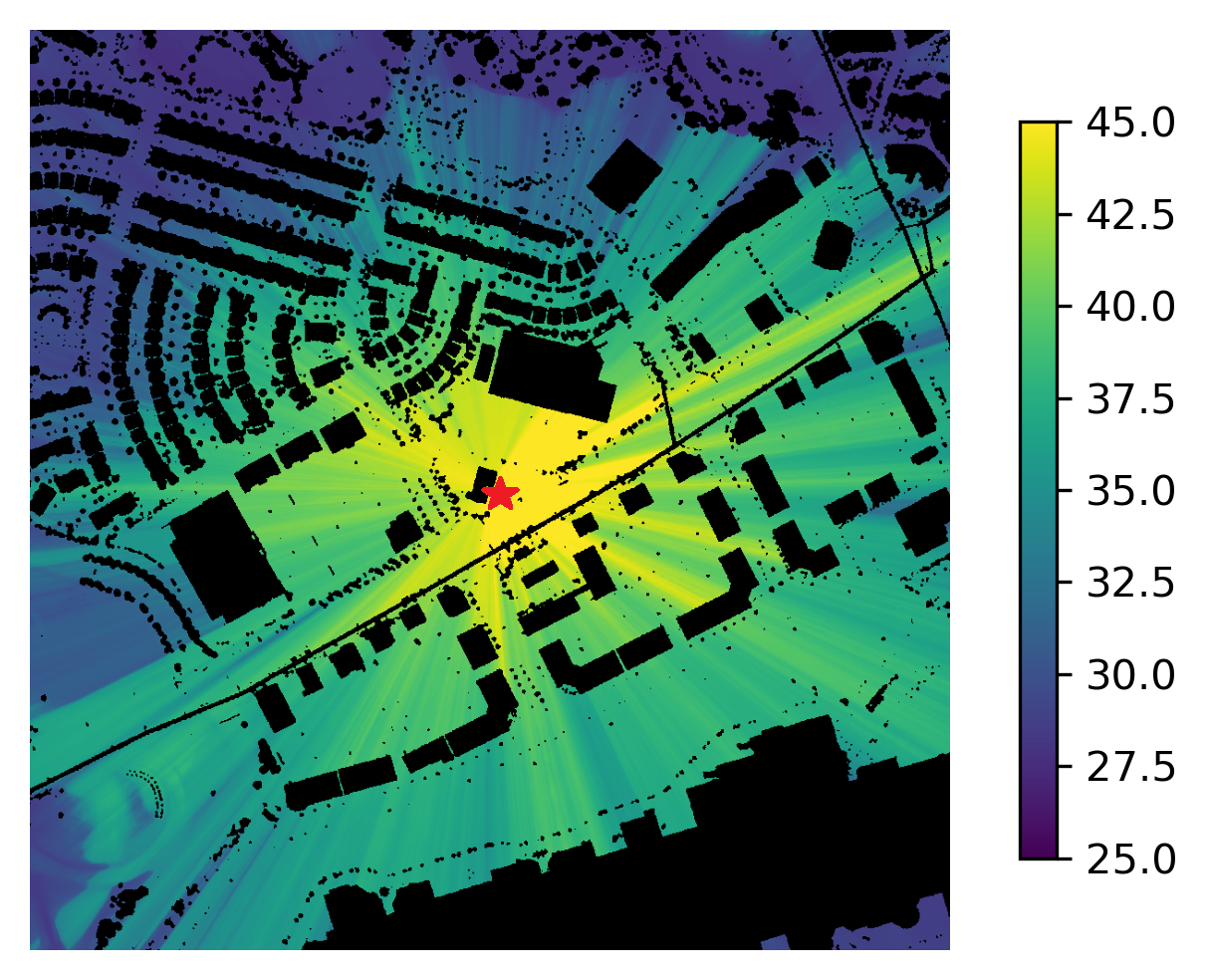}

    \caption{Illustrative heatmap showing spatially dependent 95\% confidence intervals (width ranging from 25 to 45 dB). Black regions represent areas inside buildings where path loss is not predicted, as training data is outdoor-only. Larger intervals indicate higher uncertainty, showing 20 dB variation across the map compared to a fixed 27 dB for the homoscedastic approach.}
    \label{fig:heatmap}
\end{figure}

Using the NLL-optimized shared parameter architecture model, we generate a 95\% prediction interval heatmap for a centrally located 20-m transmit tower providing coverage to a 1km x 1km urban area, shown in Fig. \ref{fig:heatmap}. The link distance and total obstruction depth features are obtained from the underlying height map, and a frequency of 3500 MHz is used. The heatmap highlights areas with large prediction intervals, indicating that higher margins are required.

The ML model from Table \ref{tab:data_table}, which has a constant prediction variance of 6.9 dB, yields a 95\% confidence interval width of 27 dB, along with the poorer calibration. The constant variance means that some predictions have overly narrow intervals, while others will be overly broad, depending on the true underlying heteroscedasticity. In contrast, a heteroscedastic approach adapts the interval width to the local uncertainty, providing narrower intervals where the model is confident and wider intervals where the predictions are less certain, resulting in more informative uncertainty estimates. Since both the heteroscedastic (NLL) and homoscedastic (MSE) models use identical features, we can be confident that the better calibration and wider intervals provided by the NLL model are better representative of reality.

\subsection{Uncertainty as Model Diagnostics}
An additional observation emerged when inspecting the predicted uncertainty in Section \ref{subsection:usecase}. Although the model was well calibrated and provides good sharpness, the highest predicted variances appeared in scenarios that are physically "easy" (e.g., LOS conditions). This behavior reflects a valuable property of data-driven uncertainty estimation. Because the training set contained very few LOS samples (see \cite{ethierFeatures,bocusPaper} for more details), the model correctly expressed higher uncertainty despite the underlying propagation conditions being less complex. The predicted variance reflected "regions where the model is ignorant" rather than "regions where the propagation is complex." Well-calibrated uncertainty estimates diagnose where training data is insufficient, enabling targeted data collection and feature development. Rather than a limitation, identifying these gaps through uncertainty analysis provides a rigorous path to enhanced model reliability and coverage.

\section{Conclusions}
\label{sec:results}
This work demonstrates that NLL is an effective loss function for optimizing ML-based path loss models. The shared-parameter architecture yielded the best training stability and test performance. Rigorous evaluation of accuracy, calibration, and sharpness confirmed the reliability of the uncertainty estimates. The prediction of path loss distributions enables more informed coverage and interference analyses than point predictions alone. Future work will incorporate additional, more varied training samples and explore new model features.

\section*{Acknowledgments}
The author would like to thank Alexis Bose and Dr. Paul Guinand for sharing their certain knowledge on uncertainty.


\end{document}